\documentclass{article}

\usepackage{arxiv}

\usepackage[utf8]{inputenc} 
\usepackage[T1]{fontenc}    
\usepackage{hyperref}       
\usepackage{url}            
\usepackage{booktabs}       
\usepackage{amsfonts}       
\usepackage{nicefrac}       
\usepackage{microtype}      
\usepackage{lipsum}
\usepackage{graphicx}
\usepackage{amsmath}
\usepackage{natbib}
\usepackage{tikz}
\usepackage{subcaption}

\usetikzlibrary{arrows,shapes,positioning,shadows,trees, 3d, calc, shapes.misc, patterns}

\tikzset{
    in_out/.style = {draw, thick, rectangle, minimum height = 0.6cm,  minimum width = 1.2cm, rounded corners=5, fill=gray!30, align=center},
    archi_layer/.style = {draw, thick, circle, align=center},
    a_dia/.style = {draw, thick, diamond, align=center},
    input/.style = {coordinate},
    num_circle/.style = {draw, thick, circle, black}}

\definecolor{input_purple}{HTML}{431cce} 
\definecolor{middle_beige}{HTML}{FAE7D6}
\definecolor{output_red}{HTML}{CE1C4E}
\definecolor{text_white}{HTML}{ECECEC}

\newcommand{\argmin}{\mathop{\mathrm{argmin}}}

\newcommand{\loss}{\ell_{\textrm{MSE}}}
\newcommand{\dataset}{\mathcal{D}}

\newcommand{\weights}{\delta}
\newcommand{\state}{\theta}

\newcommand{\ensw}{\Delta(\alpha, \lambda)}

\newcommand{\archi}{\alpha}
\newcommand{\ensarchi}{\mathcal{A}}
\newcommand{\hp}{\lambda}

\newcommand{\enshp}{\Lambda(\alpha)}

\newcommand{\Sp}{\Omega}
\newcommand{\R}{\mathbb{R}}
\newcommand{\N}{\mathbb{N}}
\newcommand{\y}{\mathbf{y}}

\title{Automated Spatio-Temporal Weather Modeling for Load Forecasting}

\author{
 Julie Keisler \\
  EDF R\&D, Lab Paris Saclay\\
  INRIA\\
  \texttt{julie.keisler@edf.fr} \\
   \And
 Margaux Br\'eg\`ere \\
  EDF R\&D, Lab Paris Saclay\\
  LPSM, Sorbonne Universit\'e\\
  \texttt{margaux.bregere@edf.fr} \\
}

\begin{document}
\maketitle
\begin{abstract}
Electricity is difficult to store, except at prohibitive cost, and therefore the balance between generation and load must be maintained at all times. Electricity is traditionally managed by anticipating demand and intermittent production (wind, solar) and matching flexible production (hydro, nuclear, coal and gas). Accurate forecasting of electricity load and renewable production is therefore essential to ensure grid performance and stability. Both are highly dependent on meteorological variables (temperature, wind, sunshine). These dependencies are complex and difficult to model. On the one hand, spatial variations do not have a uniform impact because population, industry, and wind and solar farms are not evenly distributed across the territory. On the other hand, temporal variations can have delayed effects on load (due to the thermal inertia of buildings). With access to observations from different weather stations and simulated data from meteorological models, we believe that both phenomena can be modeled together.  In today's state-of-the-art load forecasting models, the spatio-temporal modeling of the weather is fixed. In this work, we aim to take advantage of the automated representation and spatio-temporal feature extraction capabilities of deep neural networks to improve spatio-temporal weather modeling for load forecasting. We compare our deep learning-based methodology with the state-of-the-art on French national load. This methodology could also be fully adapted to forecasting renewable energy production.
\end{abstract}


\section{Introduction}
\label{sec:intro}
The cost of large-scale electricity storage remains high, and the current systems in use remain inefficient. Furthermore, the secure and smooth operation of the power grid depends on maintaining a constant and precise balance between electricity production and demand. The aforementioned equilibrium is achieved via the adaptability of programmable power plants, which modify their production in accordance with load forecasts. It is therefore essential to have accurate forecasts of both electricity demand and the output of renewable energy sources in order to schedule power plants and maintain grid stability. The two signals depend on meteorological variables, specifically temperature, wind speed, and solar radiation, which vary in both space and time.
As consumer demand and renewable energy generation facilities are not evenly distributed across a given area, variations in meteorological conditions at a particular location will affect these signals. In addition, temporal weather variations can have a delayed effect, particularly with regard to the load, due to the thermal inertia of buildings and the reactivity of consumers. It can be assumed that the manner in which temporal and spatial variations in weather patterns are modelled has a significant impact on the efficacy of the forecasting models.

This article concentrates on short-term load forecasting with a forecast time horizon of one day. Such forecasts enable power system operators to make adjustments to production and spot market prices. This signal is challenging to forecast due to its dependence on a multitude of variables, including meteorological factors (temperature, wind, etc.) and calendar-related elements (holidays, weekdays, etc.). Consequently, the models employed in the industry and those that have been successful in load forecasting competitions (see  \citet{farrokhabadi2022day}, for a recent example) are regression-based models, such as Generalized Additive Models (GAMs) or tree-based models. In general, lagged load is not employed.
While this variable offers valuable insight, it can also limit the model's interpretability by reducing the importance of other variables. Furthermore, the model would become unusable in case of data stream issues. To address this challenge while leveraging the insights offered by lagged load, a promising approach is to construct a static model only based on the explanatory variables and then recalibrate this model by adjusting certain parameters a posteriori using lagged load. For instance, \citet{goude_adaptation} proposes adaptive learning algorithms that combine additive models with a recursive least squares filter, while \citet{hal-18002} employs a Kalman filter to perform an online adaptation of the weights of their models.

Despite their impressive performance in various fields, such as computer vision and natural language processing, deep neural networks (DNNs)are still not widely used in the load forecasting community. However, recent work presents promising results for load forecasting using DNNs. In particular, \citet{keisler2024automated} proposed EnergyDragon, a deep neural network optimization framework designed for load forecasting. EnergyDragon automatically finds high-performance neural networks for the static part of load forecasting models and is able to outperform state-of-the-art regression models.
Since neural networks have demonstrated their ability to extract relevant features from data in a variety of formats, we thought it would be interesting to try them on raw spatio-temporal weather data to see if they could automatically find more relevant spatio-temporal representations than the fixed ones used in the state of the art. 
In summary, our contributions are as follows:

\begin{itemize}
    \item A DNN-based spatio-temporal weather modeling for load forecasting, which improves on the static modeling currently in use while remaining interpretable. 
    \item The integration of this weather modeling approach into the framework EnergyDragon.
    \item An application of our results to a concrete use case: the day-ahead French load forecasting over a turbulent period: sobriety during the year 2023.
\end{itemize}

We start this paper by presenting in section~\ref{part:related_work} the state of the art in short-term load forecasting: regression-based models, EnergyDragon and recalibration methods. In Section~\ref{part:methodology}, we show how to learn the actual spatio-temporal modeling approach with DNN. Section~\ref{part:ed} introduce how to incorporate the spatio-temporal weather modeling into EnergyDragon. Finally, Section~\ref{part:experiments} details our experimental results obtained on a real-world use case: the French national load forecasting. Section~\ref{part:conclusion} concludes the paper and presents further research opportunities.

\section{Related Work}\label{part:related_work}

The load signal can be explained almost entirely by a set of explanatory variables that do not include the past target data. Consequently, the majority of performing models are based on regression rather than time series techniques. Multiple Linear Regressions (MLRs) can be used to calculate the relationships between multiple variables.  However, the relationships between load and some exogenous variables are not linear, and thus, these models require the specification of functional forms for these variables. For instance, Generalized Additive Models (GAMs) employ a spline basis to model the nonlinear effects, as detailed in \citep{pierrot2011short}. These models, which are highly accurate for load forecasting, are used in industry and have been the winners of several competitions (see, for example, \citet{nedellec2014gefcom2012}).

DNNs have dominated the fields of computer vision and natural language processing for some years now. They offer the ability to process data in a variety of formats - e.g., text, images, graphs - make them particularly interesting for load forecasting, which depends on a large number of explanatory variables that may come from data sets in a variety of formats.  Recently, they have also revolutionized the field of weather forecasting, proving more effective than Numerical Weather Prediction (see for example \citep{pathak2022fourcastnet} and \citep{lam2022graphcast}).While the initial work was based on gridded reanalysis data, \citet{mcnally2024data} have shown that DNNs could also be effective in extracting spatio-temporal features directly from raw weather data. In the field of load forecasting, they are currently less widely used than multilinear regression models. However, \citet{keisler2024automated} have demonstrated that by optimizing the structure and hyperparameters of DNNs, it is possible to develop models that surpass the current state of the art. In their article, the authors optimize DNNs using the DRAGON package\footnote{\url{https://dragon-tutorial.readthedocs.io/en/latest/}} (see \citet{keisler23}). The models are represented using Directed Acyclic Graphs (DAGs). The search space is defined as $\Sp = (\ensarchi \times \{\enshp, \archi \in \ensarchi\})$, where $\ensarchi$ is the set of all considered architectures and $\enshp$ is the set of all considered hyperparameters induced by the architecture $\archi$. Each architecture $\archi \in \ensarchi$ is represented by a DAG $\Gamma$, where the nodes are the DNN layers and the edges are the connections between them. See \citet{keisler23} for more information about this search space.

Finally, this paper deals with short-term forecasting of electricity consumption. The COVID crisis and recent european energy crises have highlighted the importance of models that can rapidly adapt to new contexts. This is why research in the field has focused on different techniques for online model adaptation. These include the Kalman filter adaptation of Generalized Additive Models (GAMs), which won the post-covid electricity load forecasting competition (see \citet{farrokhabadi2022day} and \citet{de2022state}). The adaptation is done by multiplying the GAM effects vector by a linear correction. Let's have $x_t \in \R^D$ our features vector, with $D \in \N^\star$, $y_t \in \R$ the target and $\hat{y}_t \in \R$ the forecasted target. The static GAM model can be defined as:
\begin{align*}
    \hat{y}_t = \sum_{i=1}^{D} f_i(x_t).
\end{align*}
Let's have $f(x_t) = [f_i(x_t)]_{i=1}^D$ the GAM effects vector, the adaptation is done by fitting a vector $\state_t \in \R^D$ called state, such that:
\begin{align}
    &\hat{y}_t = \sum_{i=1}^D \state_{i,t}f_i(x_t) + \epsilon_t = \state_t^Tf(x_t) + \epsilon_t \\
    &\state_{t+1} = \state_t + \eta_t,
\end{align}
where $\epsilon_t \sim \mathcal{N}(0, \sigma^2)$ and $\eta_t \sim \mathcal{N}(0, Q)$, with $\sigma^2$ and $Q$ are time-invariant and assumed to be known. The algorithm achieves the estimation of the state $\state_t$ by computing its state posterior distribution as a Gaussian distribution: ${\state_t | (x_s, y_s)_{s<t} \sim \mathcal{N}(\hat{\state}_t, P_t )}$. The algorithm relies on the choice of $\sigma$ and $Q$. \citet{hal-18002} suggests an iterative grid search. In this work, we propose to apply a state vector $\state_t$ on the last layer of a DNN and to optimize $\sigma$ and $Q$ directly through the EnergyDragon framework, as any other hyperparameter. 

\section{Weather Modeling with Deep Neural Networks for load forecasting}\label{part:methodology}

In this work, we aim to forecast at each time step $t \in [1, \dots, T]$, with $T\in \N^\star$ a daily load variable $y_t \in \R^{H}$, using a features vector $x_t = (w_t, c_t) \in (\R^{H \times V \times I} \times \R^{H \times F})$, where $T$ represents the number of days in the data set and $H$ the number of time steps within a day. The features vector $x_t$ is made of two elements: $w_t$ gathering the spatio-temporal weather data and $c_t$ containing the other $F \in \N$ explanatory variables such as calendar data (e.g., months, years holidays). The vector $w_t \in \R^{H \times V \times I}$ contains the forecasts at time $t$ from different weather stations, or to a weather forecast grid, produced by, for example, a NWP model. The dimension $I \in \N^{\star}$ corresponds to the number of spatial points (i.e., the number of weather stations in the first case and the number of grid points in the second) and $V \in \N^{\star}$ to the number of weather variables present in $w_t$ (e.g., temperature, wind speed, solar radiation).

\subsection{Spatio-temporal weather modeling}

In order to be integrated into load forecasting models, spatio-temporal weather is deterministically transformed into "electrical" weather. Several functions are applied in order to reduce the information and extract what will be most useful for load forecasting. These functions have been defined with industry expertise, but are not adapted to a particular dataset or period. An example of such functions for the french load signal are given by the French Transmission System Operator called RTE\footnote{\url{https://www.services-rte.com/files/live/sites/services-rte/files/documentsLibrary/2022-04-01_REGLES_MA-RE_SECTION_2_F_3590_en}}.

\paragraph{Ponderation} The first step is to switch back from the multi-variate, spatial signal to an aggregated univariate signal. The $I$ spatial locations are not necessarily evenly distributed throughout the considered region and do not contribute equally to the electrical weather. For instance locations in densely-populated parts have more weights than others located in isolated areas. Let's denote $w^{v,i}_{h} \in \R$ the forecast of the weather variable  $v$ (e.g. temperature) at time step $h$ of the location $i$, and $a^i \in [0,1]$ the weight of the location $i$. The weights are shared accross the variables $v$. The aggregated signal at time $h$ can then be written as: 
\begin{equation}
    w^{v}_{h} = \sum_{i=1}^Ia^{i}w^{v,i}_{h} \textrm{, with: } \sum_{i=1}^{I}a^i = 1.
\end{equation}
This behavior can easily be reproduced with a Multi-Layer Perceptron (MLP):
\begin{equation}
    w^{v}_{h} = A\textbf{w}^v_{h} + b \textrm{, with } \textbf{w}^v_{h} = [w^{v,i}_{h}]_{i=1}^{I} \textrm{, } A = [a^i]_{i=1}^{I} \textrm{ and } b=0
\end{equation}
However, a Deep Neural Networks requires the scaling of the input data. Each $v$ variable is scaled independently, so that variables with large amplitudes (e.g. temperature) don't override the others. We scale each location $i$ independently and denote {$\tilde{w}^{v,i}_{h} = (w^{v,i}_{h} - \textrm{min}^{v,i})/(\textrm{max}^{v,i}  - \textrm{min}^{v,i})$} the min-max scaled version of $w^{v,i}_{h}$, with $\textrm{min}^{v,i} = \underset{h \in [1, \dots, H]}{\textrm{min }}w^{v,i}_{h} \in \R$ and $\textrm{max}^{v,i} = \underset{h \in [1, \dots, H]}{\textrm{max }}w^{v,i}_{h} \in \R$. If we consider the aggregated target to also be scaled, with $\textrm{min}^{v} = \underset{h \in [1, \dots, H]}{\textrm{min }}w^{v}_{h} \in \R$ and $\textrm{max}^{v} = \underset{h \in [1, \dots, H]}{\textrm{max }}w^{v}_{h} \in \R$, we have:
\begin{align*}
    \tilde{w}^v_{h} = \frac{w^v_{h} - \textrm{min}^v}{\textrm{max}^v - \textrm{min}^v} 
    &= (\sum_{i=1}^{I} a^iw^{v,i}_{h} - \textrm{min}^v)/(\textrm{max}^v - \textrm{min}^v)\\ 
    &= \big(\sum_{i=1}^{I} \tilde{w}_{h}^{v,i}(\textrm{max}^{v,i} - \textrm{min}^{v,i}) + \textrm{min}^{v,i} - \textrm{min}^v\big) / (\textrm{max}^v - \textrm{min}^v)\\
    &= \sum_{i=1}^{I} a^i\frac{\textrm{max}^{v,i} - \textrm{min}^{v,i}}{\textrm{max}^{v} - \textrm{min}^{v}} \tilde{w}_{h}^{v,i} + \sum_{i=1}^I \frac{\textrm{min}^{v,i}-\textrm{min}^v}{\textrm{max}^{v}-\textrm{min}^v}\\
    &= A_v \tilde{\textbf{w}}_{h}^v + b_v
\end{align*},

with $\tilde{\textbf{w}}_{h}^v = [\tilde{w}_{h}^{v,i}]_{i=1}^{I}$, $A_v = [a^i(\textrm{max}^{v,i} - \textrm{min}^{v,i})/(\textrm{max}^{v} - \textrm{min}^{v})]_{i=1}^I$ and $b_v = \sum_{i=1}^I(\textrm{min}^{v,i}-\textrm{min}^v)/(\textrm{max}^{v}-\textrm{min}^v)$. Therefore, we need $V$ MLP layers to aggregate the data variable by variable.

\subsection{Temperature smoothing}

Load does not respond instantaneously to changes in the weather. In particular, temperature effects are more gradual due to the thermal inertia of buildings. This is why the concept of smoothed temperature is useful for understanding the factors that influence electricity consumption. Exponential smoothing is typically employed in this context. We denote, for a day $t$, $T_t = [T_{t,1}, \dots, T_{t,H}]\in \R^{H}$ the aggregated temperature and $\overline{T}_t = [\overline{T}_{t,1}, \dots, \overline{T}_{t,H}]\in \R^{H}$ the smoothed version. We define:
\begin{equation}\label{eq:es}
    \overline{T}_{t,1} = (1-\alpha) T_{t,1} + \alpha \overline{T}_{t-1, H} \textrm{, and, } \forall i \in [2, H]: \overline{T}_{t,i} = (1-\alpha) T_{t,i} + \alpha \overline{T}_{t, i-1},
\end{equation}
where $\alpha \in [0,1]$ is the smooth coefficient, which can be optimized.

\paragraph{Recurrent Neural Networks} Smoothed temperature requires at each time step $t$ to pass $\overline{T}_{t,H} \in \R$ to the next time step $t+1$. Such information passing can be reproduced by Recurrent Neural Networks (RNNs), which are designed with a memory vector. The equations of a recurrent neural network with input $T_t \in \R^H$ and output $\overline{T}_t \in \R^H$ can be written as:
\begin{equation*}
    \overline{T}_{t} = \phi(T_t W_1^T + b_1 + \overline{T}_{t-1} W_2^T + b_2)\,,
\end{equation*}
where $\phi$ is an activation function (typically non-linear), and $W_{1}\in \R^{H \times H}$, $W_{2} \in \R^{H \times H}$, $b_{1} \in \R^H$ and $b_{2} \in \R^H$ are some parameters which can be learned through gradient descent. For writing simplicity, we now index our temperature by $t^\star$, such that, if $t^\star=\{t, i\}$, we have, if $i<H$:
\begin{align*}
    \left\{
    \begin{array}{ll}
        &t^\star= \{t, i+1\}\\
        &\textrm{else, } t^\star+1 = \{t+1, 1\}.    \end{array}
\right.
\end{align*}
Let $\tau > 0$.
To compute $\overline{T}_{t^\star}$ based on the smoothed temperature at instant $t^\star-\tau$, $\overline{T}_{t^\star-\tau}$, and the sequence of temperatures $T_{t^\star-\tau +1},\dots,T_{t^\star}$, we have:
\begin{align}
\overline{T}_{t^\star-\tau +1} &= \overline{T}_{t^\star-\tau} \nonumber\\
\overline{T}_{t^\star-\tau + 1} &=  (1-\alpha) T_{t^\star-\tau + 1} + \alpha \overline{T}_{t^\star-\tau} \nonumber\\
\vdots \quad  &= \quad  \vdots \nonumber\\
\overline{T}_{t^\star} &=   \sum_{s=0}^{\tau-1} \alpha^{s}(1-\alpha)T_{t^\star-s} + \alpha^\tau \overline{T}_{t^\star-\tau}\,.\label{eq:es_tau}
\end{align}
Based on Equation~\ref{eq:es_tau}, by setting: 
\begin{align*}
    &W_{1}=\begin{bmatrix}
        (1-\alpha) & 0 & 0  & \dots  & 0\\
        \alpha(1-\alpha) & (1-\alpha) & 0 &  \dots & 0 \\
        \vdots & \vdots & \vdots & \vdots & \vdots \\
        \alpha^{H-1}(1-\alpha) & \alpha^{H - 2}(1-\alpha) & \alpha^{H-3}(1-\alpha) & \dots & (1-\alpha)
    \end{bmatrix},\\
    &W_{2} =\begin{bmatrix}
        0  & \dots &0 & \alpha\\
        0 & \dots & 0 & \alpha^2 \\
        \vdots & \vdots & \vdots & \vdots \\
        0 & \vdots & 0 & \alpha^H
    \end{bmatrix},\\
    &\textrm{and } b_{1} = b_{2} = 0,
\end{align*}
It is possible to induce a recurrent neural network (RNN) to learn the behaviour defined by the exponential smoothing model, as set out in Equation~\ref{eq:es}. Nevertheless, this configuration results in the network optimizing a total of $\mathcal{O}(H^2)$  parameters. In the context of seeking novel approaches to temperature smoothing, the utilization of a recurrent neural network (RNN) is a logical choice. On the other hand, if the objective is to restrict the network to exponential smoothing, with optimization limited to the $\alpha$ smoothing coefficient, the necessity for optimizing a vast number of parameters renders the process complex and may ultimately prove inefficient. For this reason, we propose a new DNN layer that enables the efficient computation of one or more exponential smoothings over several batches, with only the smoothing coefficients as parameters to optimize.
\paragraph{Exponential Smoothing Layer} Let's consider a batched input of size $B \in \N^\star$, containing the temperature from the days $t-B$ to $t$: ${\textbf{T}_{t-B:t}=[T_{t-B}, \dots, T_t]\in \R^{B \times H}}$. The exponential smoothing layer first reshape this data into a size $BH$, to treat all sequences at once. We then use Equation~\ref{eq:es_tau}, with $\tau = H \times B -1$ to compute $\overline{\textbf{T}}_{t-B:t}$:
\begin{equation*}
  \begin{bmatrix} \overline{T}_{t^\star-\tau}\\
  \overline{T}_{t^\star-\tau+1} \\
  \vdots \\
  \overline{T}_{t} \end{bmatrix}
  = \begin{bmatrix}
        1 & 0 & 0  & \dots  & 0\\
        \alpha & 1 & 0 &  \dots & 0 \\
        \vdots & \vdots & \vdots & \vdots & \vdots \\
        \alpha^{\tau} & \alpha^{\tau - 1} & \alpha^{\tau-2} & \dots & 1
    \end{bmatrix} 
    \begin{bmatrix} \overline{T}_{t^\star-\tau}\\
 (1-\alpha) T_{t^\star-\tau+1} \\
  \vdots \\
 (1-\alpha) T_{t^\star} \end{bmatrix} \,,
\end{equation*}

\begin{equation}
    \overline{\mathbf{T}}_{t^\star-\tau:t^\star} = M \times \big[\overline{T}_{t^\star-\tau} \, | \, (1-\alpha)\mathbf{T}_{t^\star-\tau+1:t^\star} \big]\textrm{, with},\forall i \geq j: M_{i,j} = \alpha^{i-j}\,.
\end{equation}
Finally, we reshape $\overline{\mathbf{T}}_{t^\star-\tau:t^\star} \in \R^{HB}$ back to the original shape $\overline{\mathbf{T}}_{t-B:t}\in \R^{H\times B}$. The matrix $M$ is constructed during the forward pass as its shape and formula depend on the batch size $B$. Given that the coefficient $\alpha$ belongs to $[0,1]$, it is encoded through a sigmoid: $\alpha = \textrm{Sigmoid}(\overline{\alpha})=1/(1 + \textrm{exp}(- \overline{\alpha})) \in [0,1] \textrm{ for } \overline{\alpha} \in \R$, where $\overline{\alpha}$ would be the weight optimized through back-propagation.

In our search space we let the optimization framework choose between the Exponential Smoothing Layer, a RNN layer, a Long-Short Term Memory (LSTM) layer and a Gated Recurrent Unit (GRU) layer, to perform the smoothing operation (see Section~\ref{part:sp}).

\subsection{Online learning}
The last layer of the search space from \citet{keisler2024automated} is a linear layer, transforming an input $h_t \in \R^{H \times D}$ into an output $y_t \in \R^{H}$, where $y_t$ is the load consumption for the day $t$, $H$ the number of instants during the day and $D \in \N^\star$ is the hidden dimension within the network before the last layer. Let's name $A_F \in \R^D$ and $b_F \in \R$ respectively the weights and bias matrices of this last MLP layer, we have:
\begin{equation}\label{MLP_kalman}
    y_t = A_F h_t + b_F = \sum_{i=1}^{D} a_F^i h_t^i + b_F.
\end{equation}
To adapt our DNN, we use a daily Kalman state vector $\state_t \in \R^{D}$ to adapt the coefficients of equation~\ref{MLP_kalman}:
\begin{align}
    &\tilde{y}_t = \state_t^T(A_F h_t + b_F) = \sum_{i=1}^{D} \state_t^i(a_F^i h_t^i + b_F) + \epsilon_t \label{eq:kalmann1}\\
    &\state_{t+1} = \state_t + \eta_t\label{eq:kalmann2},
\end{align}
where $\epsilon_t \sim \mathcal{N}(0, \sigma^2)$ and $\eta_t \sim \mathcal{N}(0, Q)$. \citet{hal-18002} suggests to use iterative grid search for $\sigma \in \R$ the diagonal coefficients of $Q \in \R^{D \times D}$. This search can be quite expensive, with a complexity of $\mathcal{O}(LD^2)$, where $L$ is number of values that the coefficients of $Q$ may take. We experimented empirically that the number of coefficients of the last MLP layer is usually larger than the number of coefficient of the GAMs. The iterative grid search was not usuable in this case, therefore we included the optimization of $\sigma$ and the coefficients of $Q$ within EnergyDragon search space (see Section~\ref{part:sp}.

\section{Automated weather modeling}\label{part:ed}

This Section presents the integration of the various elements presented in Section~\ref{part:methodology}, namely the weather modeling and Kalman adaptation modules, into EnergyDragon with the objective of optimizing them.

\subsection{Objective function}
\label{sec:obj}

The objective is to identify the optimal DNN $\hat{f} \in \Sp$ with the lowest forecast error on a given load signal with a short forecast horizon (e.g., 24 hours). The load dataset, denoted by $\dataset$, contains the spatio-temporal $W$ data, the explanatory variables $C$ and the target (the load signal) $Y$. For any subset $\dataset_0 = ((W_0, C_0), Y_0)$, the forecast error $\loss$ is defined as:
\begin{align*}
  \loss \colon & \Sp \times \dataset \to \R \\
  &f \times \dataset_0 \mapsto \loss\big(f(\dataset_0)\big) = \loss\big(Y_0, f(W_0, C_0)\big) = \textrm{MSE}\big(Y_0, f(W_0, C_0)\big) \,.
\end{align*}
Where MSE is the Mean Squared Error. Each DNN $f \in \Sp$ is parameterized by:
\begin{itemize}
    \item $\archi \in \ensarchi$, its architecture, optimized by the framework.
    \item $\hp \in \enshp$, its hyperparameters, optimized by the framework, where $\enshp$ is induced by $\archi$. The hyperparameters include $Q$ and $\sigma$ from the Kalman adaptation. It should be noted that the shape of $Q$ depends on the architecture and hyperparameters of the networks.
    \item $\weights \in \ensw$, the DNN weights, where $\ensw$ is generated by $\archi$ and $\hp$ and optimized by gradient descent when training the model.
\end{itemize}
The optimization process is done in several steps. First, the optimal DNN weights $\hat{\weights} \in \ensw$ for a given architecture $\alpha \in \ensarchi$ and set of hyperparameters $\lambda \in \enshp$ are found using gradient descent over the training set $\dataset_{\textrm{train}} = \big((W_{\textrm{train}}, C_{\textrm{train}}), Y_{\textrm{train}})\big)=(X_{\textrm{train}}, Y_{\textrm{train}})$:
\begin{align*}
    \hat{\weights} \in \underset{\weights \in \ensw}{\argmin}\Big(\loss\big(f_{\weights}^{\archi, \hp}(X_{\textrm{train}},Y_{\textrm{train}})\big) \Big)\,.
\end{align*} 
Once the DNN is trained, the performance of the selected $\archi$ and $\hp$ are evaluated on $\dataset_{\textrm{valid}}$. As $Q$ and $\sigma$ are part of $\lambda$, the evaluation is done using the- Kalman recalibration of the model First, the state vector $\state \in \R^{T\times D}$ is estimated on the last MLP layer of the trained DNN $f^{\alpha, \lambda}_{\hat{\weights}}$ using Equations~\ref{eq:kalmann1} and \ref{eq:kalmann2}. Let's have $\Theta_\lambda\big(f_{\hat{\weights}}^{\alpha, \lambda}(X_{\textrm{valid}})\big)$ the recalibration of $f_{\hat{\weights}}^{\alpha, \lambda}(X_{\textrm{valid}})$ as defined Equation~\ref{eq:kalmann1}. The architecture $\archi$ and hyperparameters $\hp$ are optimized as:
\begin{align*}
    (\hat{\archi}, \hat{\hp}) \in \underset{(\archi, \lambda) \in (\ensarchi \times \enshp)} \argmin\Bigg(\loss\Big(\Theta_\lambda\big(f_{\hat{\weights}}^{\alpha, \lambda}(X_{\textrm{valid}})\big), Y_{\textrm{valid}}\Big)\Bigg) \,.
\end{align*}

The framework output will be $\ell_{MAPE}$, the Mean Absolute Percentage Error. Given a load series $Y = (\y_1 \dots \y_n)$ and the predictions $\hat{Y} = (\hat{\y}_1, \dots \hat{y}_n)$, $\textrm{MAPE}(Y, \hat{Y}) = 1/n \sum_{i=1}^{n}\big|(\y_i - \hat{\y}_i)/\y_i\big| \, $. The MAPE is computed using the DNN with the best architecture, hyperparameters, weights and calibration using Kalman on the test dataset:

\begin{align*}
    \ell_{MAPE}\Big(\Theta_{\hat{\lambda}} \big( f_{\hat{\weights}}^{\hat{\archi}, \hat{\hp}}(X_{\textrm{test}}) \big),Y_{\textrm{test}}\Big) \,.
\end{align*}
In the following section (\ref{part:sp}), we explicit our search space, defined by $\ensarchi$ and $\enshp$.

\subsection{Search space}\label{part:sp}

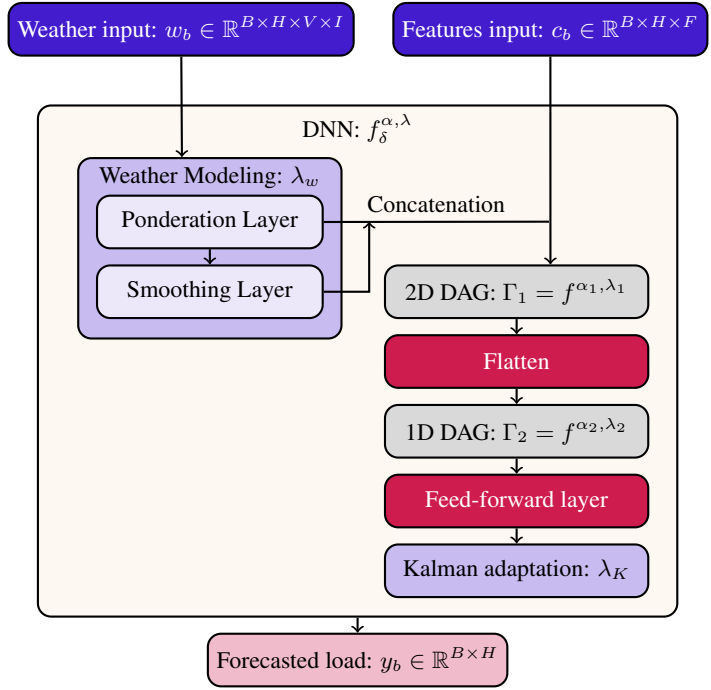
\begin{figure}[htpb]
    \centering
    \begin{tikzpicture}[auto, thick, font=\footnotesize]
        \draw
        node [in_out, minimum height = 0.7cm, minimum width=4cm, fill=input_purple, text=white](input_w) {Weather input: $w_b \in \R^{B \times H \times V \times I}$}
        
        node [in_out, minimum height = 0.7cm, minimum width=4cm, fill=input_purple, text=white, right=0.5cm of input_w](input) {Features input: $c_b \in \R^{B \times H \times F}$}
        
        node [in_out, minimum height = 6.8cm, minimum width=8.5cm, below left=1cm and 0.4cm of input.east, fill=middle_beige!30, label={[align=center, anchor=north]north:DNN: $f^{\alpha, \lambda}_{\weights}$}] (dnn) {}
        node [in_out, minimum height = 2.4cm, minimum width=3.5cm ,  fill=input_purple!30, below left=0.7cm and 0.2cm of dnn.north, label={[align=center, anchor=north]north:Weather Modeling: $\hp_w$}] (wm) {}
        node [in_out, fill=input_purple!10, minimum height = 0.7cm, minimum width=3cm , below=0.5cm of wm.north] (ponderation) {Ponderation Layer}
        node [in_out, fill=input_purple!10, minimum height = 0.7cm, minimum width=3cm , below=0.2cm of ponderation] (smoothing) {Smoothing Layer}
        node [in_out, minimum height = 0.7cm, minimum width=3.5cm , right= 0.8cm of smoothing.east] (dag1) {2D DAG: $\Gamma_1 = f^{\archi_1, \hp_1}$}
         node [in_out, minimum height = 0.7cm, minimum width=3.5cm, below=0.2cm of dag1, fill=output_red, text=white] (mlp1) {Flatten}
        node [in_out, minimum height = 0.7cm, minimum width=3.5cm , below=0.2cm of mlp1] (dag2) {1D DAG: $\Gamma_2 = f^{\archi_2, \hp_2}$}
         node [in_out, minimum height = 0.7cm, minimum width=3.5cm, below=0.2cm of dag2, fill=output_red, text=white] (mlp2) {Feed-forward layer}
         node [in_out, minimum height = 0.7cm, minimum width=3.5cm, fill=input_purple!30, below=0.2cm of mlp2] (kalman) {Kalman adaptation: $\lambda_K$}
         node [in_out, minimum height = 0.7cm, minimum width=3.5cm, below=0.2cm of dnn, fill=output_red!30] (out) {Forecasted load: $y_b \in \R^{B \times H}$};
        \draw[->](input_w) -- (wm.107); 
        \draw[->](input) -- (dag1.039); 
        \draw[-] (ponderation.east) -- node[above]{Concatenation} (4.89, -2.56);
        \draw[->](ponderation) -- (smoothing);
        \draw[->](smoothing.east) -| (2.5, -2.56);
        \draw[->](dag1) -- (mlp1);
        \draw[->](mlp1) -- (dag2);
        \draw[->](dag2) -- (mlp2);
        \draw[->](mlp2) -- (kalman);
        \draw[->](dnn) -- (out);
    \end{tikzpicture}
    \caption{Daily meta-model for load datasets from \citet{keisler2024automated}, with the integration of the weather modeling and the Kalman adaptation modules.}
    \label{fig:metamodel_load}
\end{figure}

The search space used in this work extends the one from \citet{keisler2024automated} by adding a weather modeling and a Kalman module, and is depicted Figure~\ref{fig:metamodel_load}. Each DNN $f \in \Sp$ maps two batched inputs: $w_b \in \R^{B \times H \times V \times I}$ containing the spatio-temporal weather and $c_b \in \R^{B \times H \times F}$ containing the other explanatory variables into a target $y_b \in \R^{B \times H}$, where $B \in \N^\star$ represents the size of the batch.
\paragraph{Weather Modeling} Tdesignated as $w_b$, is initially processed by a Weather Modeling module containing $V$ ponderation layers and a smoothing layer as defined Section~\ref{part:methodology}. Each of the $V$ weighting layers, designated as $v$ is associated with an MLP layer, enabling the $I$ signals to be weighted into $F_v$ signals, with $I >> F_v$. This enables the network to identify multiple potential weightings. The $F_W = \sum_{v=1}^V F_v$ weighted signals are then concatenated into a vector of size $B \times H \times F_W$. The $F_T$ signals corresponding to aggregated temperatures are smoothed by a smoothing layer being either a recurrent network (RNN, LSTM or GRU), or an exponential smoothing layer as defined in Section~\ref{part:methodology}. They are then concatenated to the vector, now of size $B \times H \times (F_w + F_T)$. The set of hyperparameters for the Weather Modeling module is called $\hp_w$ and include the $V$ output dimensions of the weighting layers for each of the $V$ variables, as well as the type of layer used for smoothing along, with the hyperparameters associated with that layer.
\paragraph{Load Forecasting Network} The vector generated by the Weather Modeling module is merged with the other vector of features, designated as $c_b$, resulting in a vector $x_b$ of size $B \times H \times (F_W + F_T + F)$, which is then fed to the load forecasting model. This model is identical to the one used by \cite{keisler2024automated}. The load forecasting network should map $x_b \in \R^{B \times H \times (F_W + F_T + F)}$ into the target $y_b \in \R^{B \times H}$. For this, two DAGs $\Gamma_1$ and $\Gamma_2$ are used. 
The graph $\Gamma_1$ is made of 2-dimensional layers operations to treat the 2-dimensional input and is parameterized by $\archi_1$ and $\hp_1$. A flattened layer follows $\Gamma_1$ to transform the 2-dimensional latent representation into a 1-dimensional one. The graph $\Gamma_2$ is then made of 1-dimensional layers operations and is parameterized by $\archi_2$ and $\hp_2$. We have $\archi = [\archi_1, \archi_2]$ and $\hp = [\hp_1, \hp_2]$. A final output layer maps the output shape of $\Gamma_2$ to $H$.

Finally, a Kalman adaptation is made using two last hyperparameters $Q$ and $\sigma$. We call $\hp_K$ this hyperparameter set. To summarize, our search space can be written as: {$(\archi, \hp) = (\{\archi_1, \archi_2\}, \{\hp_w, \hp_1, \hp_2, \hp_K\}) \in \big(\ensarchi \times \enshp\big)$}.

\section{Experiments}\label{part:experiments}

In this section, we evaluate the efficacy of our weather modeling and Kalman adaptation techniques on the French load dataset from January 2023 to May 2024. In contrast with the paper by \citet{keisler2024automated}, which compares data from a relatively stable and distant period, our analysis focuses on a more dynamic and operational context. The training period spans from 2018 to 2022 and encompasses both the global pandemic caused by the SARS-CoV-2 virus and the subsequent energy crisis at the end of 2022. The test period encompasses the winter of 2023, during which consumers were encouraged to voluntarily reduce their consumption, a period commonly referred to as the ``sobriety period''.

\subsection{Data}

\begin{figure}[htbp]
    \centering
    \includegraphics[width=0.5\linewidth]{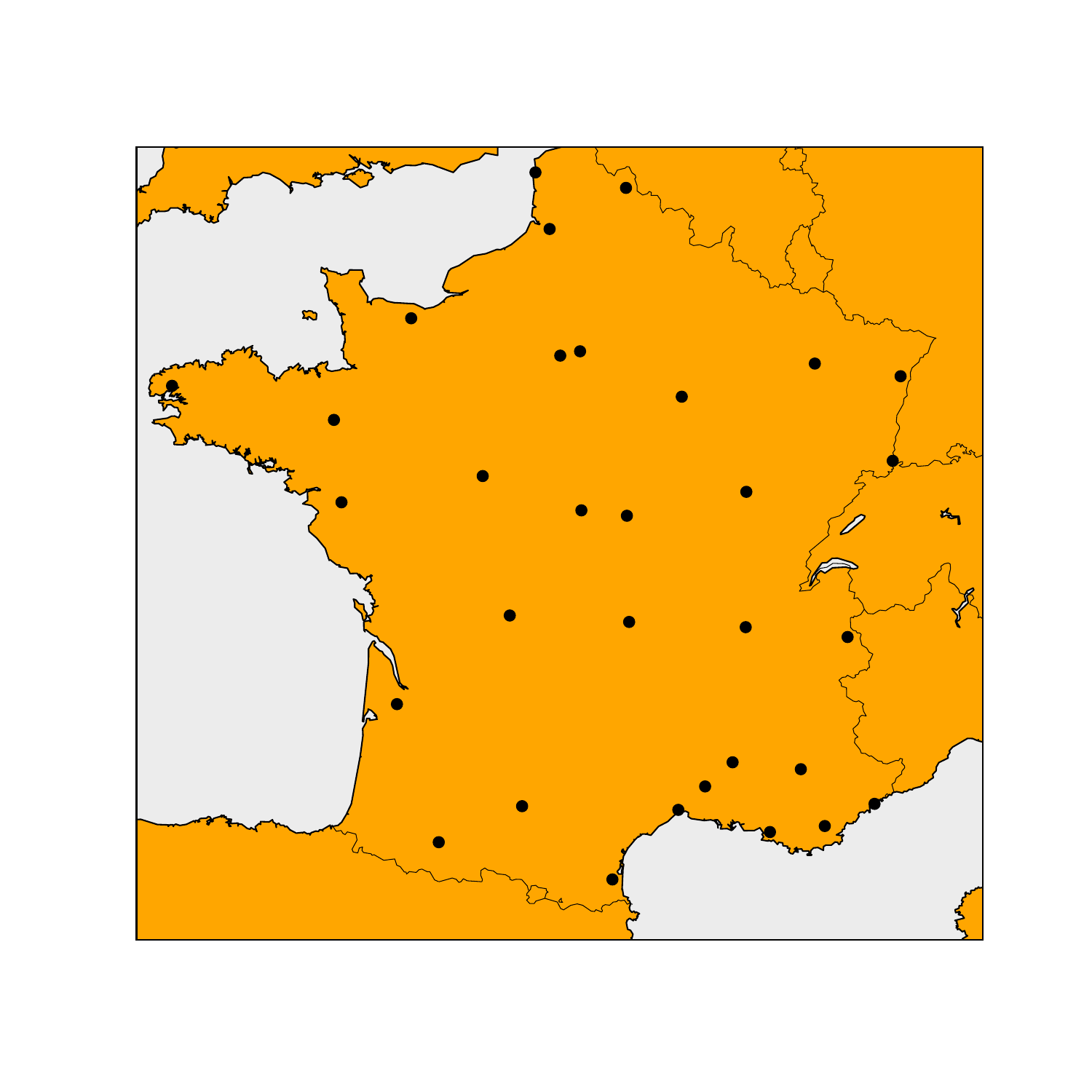}
    \vspace{-1cm}
    \caption{Location of the 32 french weather stations from our spatio-temporal weather dataset.}
    \label{fig:weather_stations}
\end{figure}

The load dataset was obtained from the website of the French Transmission System Operator (RTE)\footnote{\url{https://www.rte-france.com/eco2mix}} and contains the French national load data at half-hourly intervals. Therefore, each day contains $H=48$ time steps. The models were trained from January 2018 to December 2022 and subsequently evaluated from January 2023 to May 2024 using the MAPE. The weather data set comprises three-hourly weather forecasts for 32 weather stations across France (see Figure~\ref{fig:weather_stations}). These forecasts are provided by Meteo France \footnote{\url{https://www.data.gouv.fr/fr/organizations/meteo-france/}}. Prior to employing this data in our forecasting models, we performed a temporal linear interpolation. The other explanatory features used to explain the load data are calendar features
including the day of the week, the month, the year, and whether the day in question fell on a public holiday or a surrounding day.

\subsection{Baseline}

We compare our results to models at the state-of-the-art in load forecasting: the day-ahead load forecast provided on RTE website, a Generalized Additive Model (GAM) used in the industry and EnergyDragon as proposed by \cite{keisler2024automated}. In the case of the GAM model, a single model is calibrated for each instant, resulting in a total of 48 models. The training set was modified by removing periods corresponding to lockdowns that were implemented during the pandemic caused by the novel coronavirus. EnergyDragon produces daily forecasts of $H=48$ values, necessitating the use of a single model for all instants. A ``Covid'' feature has been incorporated into the model to indicate which periods corresponding to lockdowns are retained in the training set. With the exception of the ``Covid'' variable, the features are identical between the GAM and EnergyDragon models. The weather variables utilized in this study correspond to the weather at the 32 stations, with the data weighted and smoothed in accordance with the recommendations outlined in the RTE report on incorporating climatic contingencies into consumption forecasts\footnote{\url{https://www.services-rte.com/files/live/sites/services-rte/files/documentsLibrary/2022-04-01_REGLES_MA-RE_SECTION_2_F_3590_en}}. Both models are recalibrated in an identical manner, utilising a Kalman filter that is updated on a daily basis with data from two days ago. To optimize $Q$ and $\sigma$ for the GAM model, an Iterative Grid Search was employed with the years 2020 to 2022 as validation set. For EnergyDragon, the years 2018 to 2020 were used as training dataset ($\dataset_{\textrm{train}}$) and the years 2021 and 2022 as validation set ($\dataset_{\textrm{valid}}$). Concerning the RTE model, no information is given on the structure of the model or its recalibration.
For our model, later called ED Weather Modeling (for EnergyDragon Weather Modeling), which includes space-time weather modeling, we remove all weather-related features from EnergyDragon (all weather variables and their smoothed versions).

\subsection{Results}
\begin{table}[h!]
\vskip 0.15in

    \centering
    \begin{tabular}{lc c c}
    \toprule
        Model & MAPE & Recalibration & MAPE\\
        \midrule
        RTE & - & Not specified & 2.316 \\
        GAM & 7.429 & Kalman & 2.019 \\
        \midrule
        EnergyDragon & \textbf{2.988} & Kalman & 1.947 \\
        ED Weather Modeling & 3.501 & Kalman & \textbf{1.848}\\
        \bottomrule
    \end{tabular}
    \vskip 0.1in
    \caption{Results over 2023 - May 2024}
    \label{tab:res_rte}
\end{table}
We evaluated each algorithm from the baseline on the French load signal. Both versions from EnergyDragon were run using 20 GPUs V100. The search algorithm used for the optimisation is the steady-state evolutionary algorithm used in \citep{keisler2024automated}. The initial population is of size 100. Each algorithm was run with a global seed of 0 to ensure reproducibility. The results can be found in Table~\ref{tab:res_rte} and support the findings of \citep{keisler2024automated}. Indeed, even during an erratic period, EnergyDragon managed to beat the static version of GAM. As anticipated, GAM's Kalman recalibration is superior to EnergyDragon's static version, thereby validating the incorporation of a DNN recalibration brick. This addition, in combination with the optimization of $\sigma$ and $Q$, have proven to be effective, as evidenced by the superior performance of the recalibrated EnergyDragon model in comparison to both RTE and GAM-Kalman. With regard to the incorporation of the weather modeling module, in the recalibrated version it enables us to achieve a slight improvement (5\%) in mean absolute percentage error (MAPE) compared to EnergyDragon. 

\begin{figure}[htpb]
    \centering
    \includegraphics[width=0.8\linewidth]{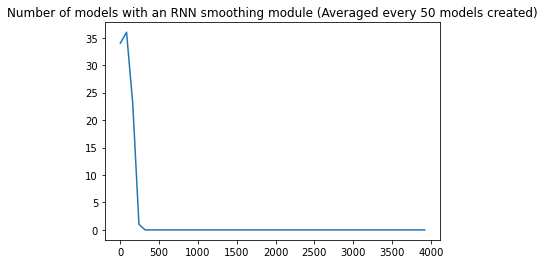}
    \caption{Number of DNNs created having a Recurrent Neural Network as smoothing layer through time.}
    \label{fig:rnn_through_time}
\end{figure}

\begin{figure}[htpb]
     \centering
     \begin{subfigure}[b]{0.95\textwidth}
         \centering
         \includegraphics[width=\textwidth]{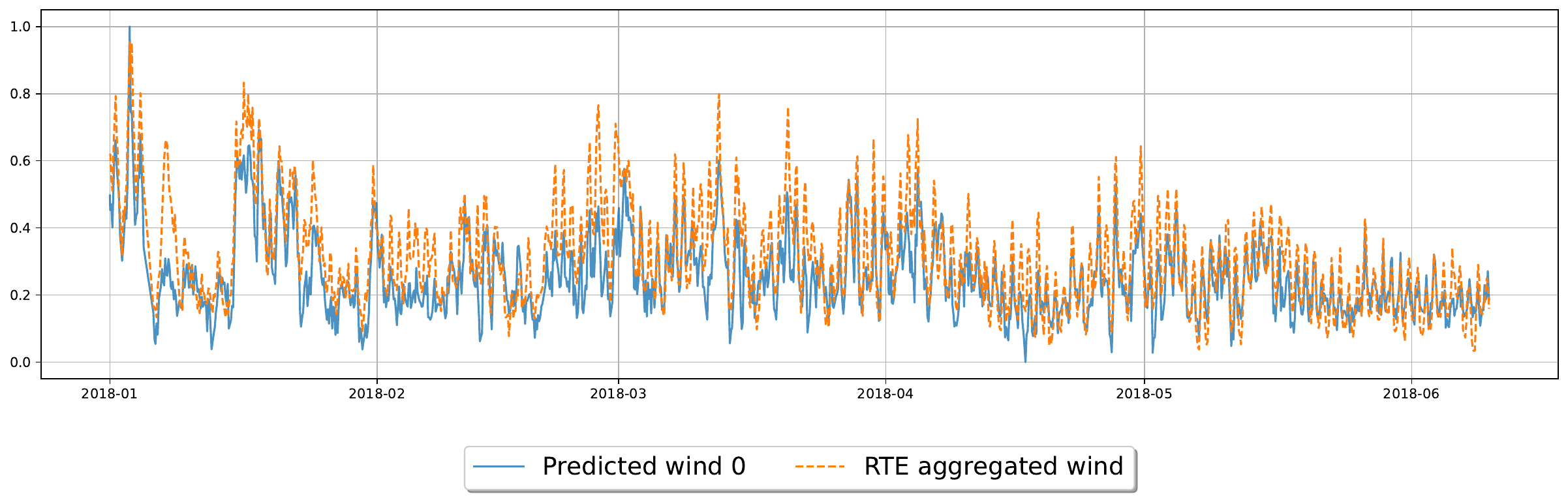}
         \caption{Dotted line: the wind signal aggregated using weights from RTE, used in the GAM and EnergyDragon models. Solid line: the wind signal aggregated by the weather modeling module.}
         \label{fig:wind}
     \end{subfigure}
     \hfill
     \begin{subfigure}[b]{0.95\textwidth}
         \centering
         \includegraphics[width=\textwidth]{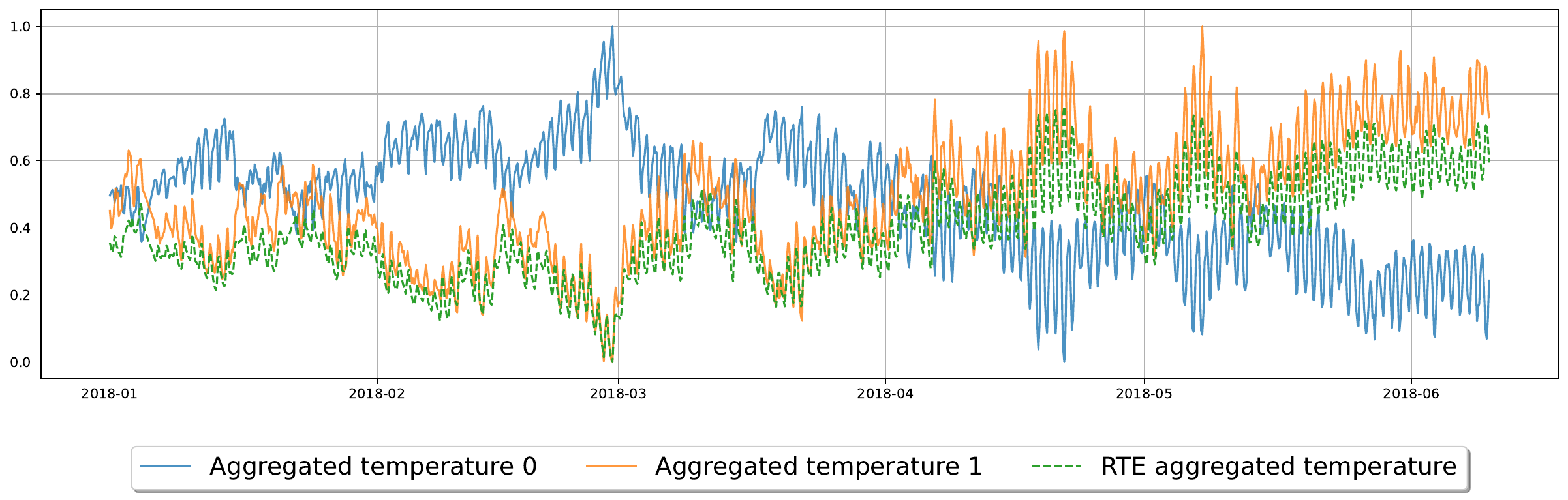}
         \caption{Dotted line: the temperature signal aggregated using weights from RTE, used in the GAM and EnergyDragon models. Solid lines: two aggregated temperature signals found by the weather modeling module.}
         \label{fig:temp}
     \end{subfigure}
     \hfill
     \begin{subfigure}[b]{0.95\textwidth}
         \centering
         \includegraphics[width=\textwidth]{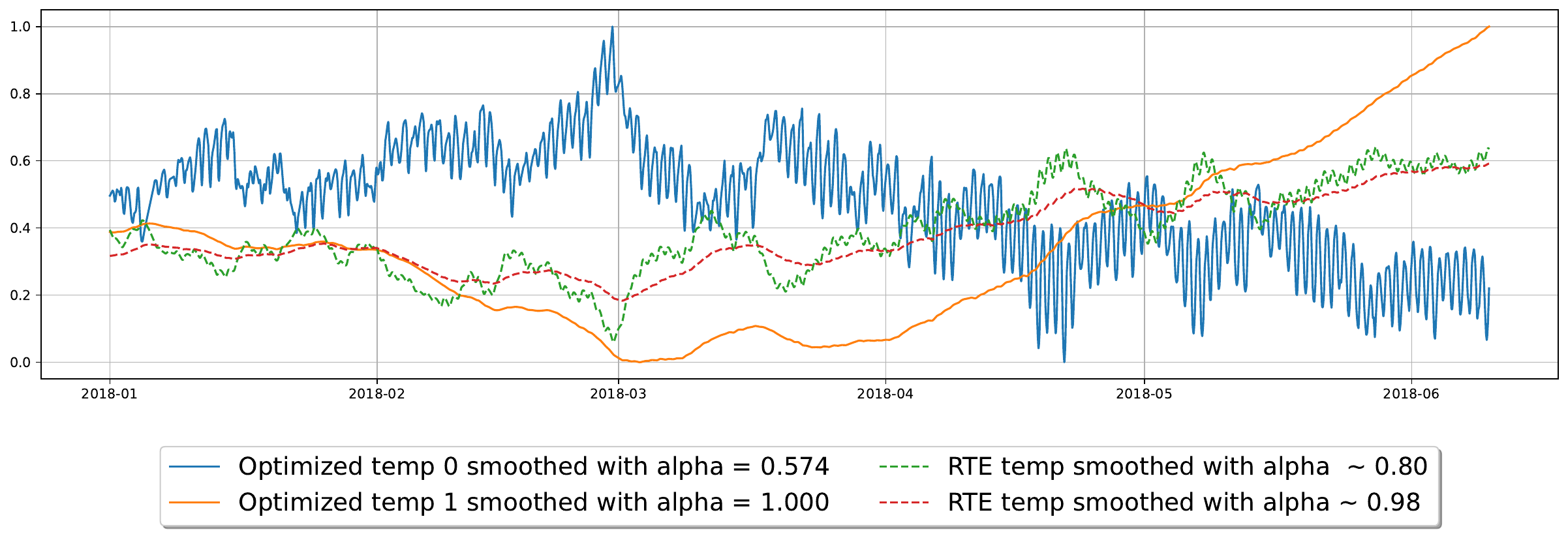}
         \caption{Dotted line: the smoothed temperature signals used in the GAM and EnergyDragon models. Solid lines: the two aggregated temperature signals smoothed by the weather modeling modeling module.}
         \label{fig:smoothed_temp}
     \end{subfigure}
        \caption{Comparison between the weather as modeled within the GAM and EnergyDragon models, versus the weather modeled by the DNN based Weather Modeling module, for a model having a MAPE of 2.1\%}
        \label{fig:wm_1}
\end{figure}

\begin{figure}[htpb]
     \centering
     \begin{subfigure}[b]{0.95\textwidth}
         \centering
         \includegraphics[width=\textwidth]{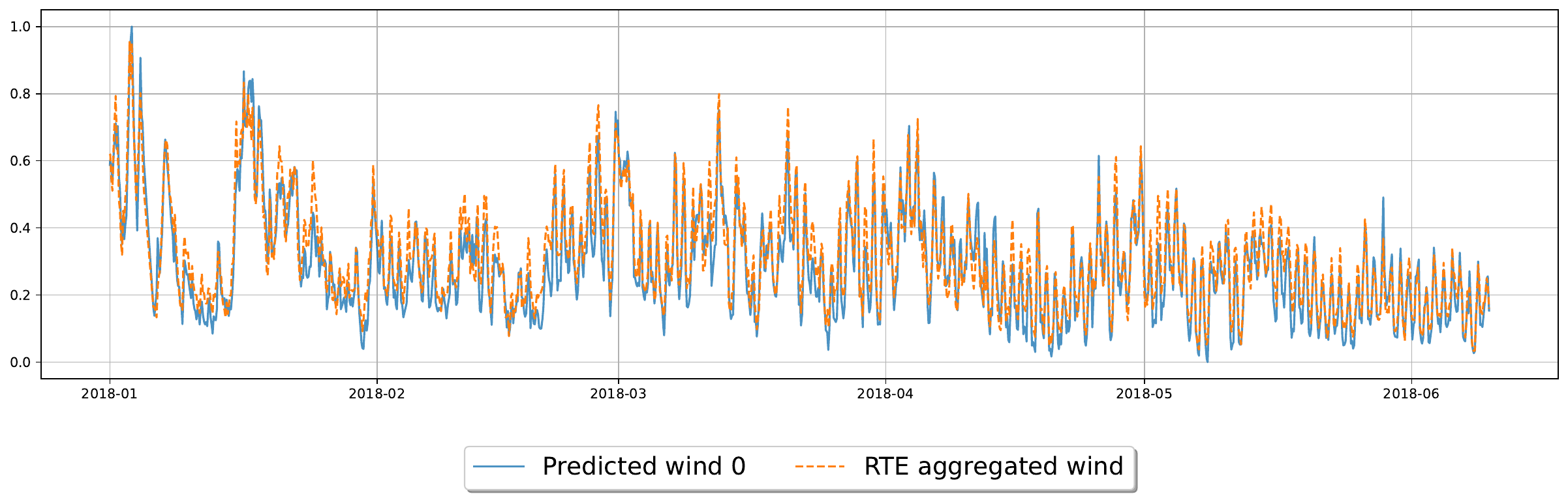}
         \caption{Dotted line: the wind signal aggregated using weights from RTE, used in the GAM and EnergyDragon models. Solid line: the wind signal aggregated by the weather modeling module.}
         \label{fig:wind2}
     \end{subfigure}
     \hfill
     \begin{subfigure}[b]{0.95\textwidth}
         \centering
         \includegraphics[width=\textwidth]{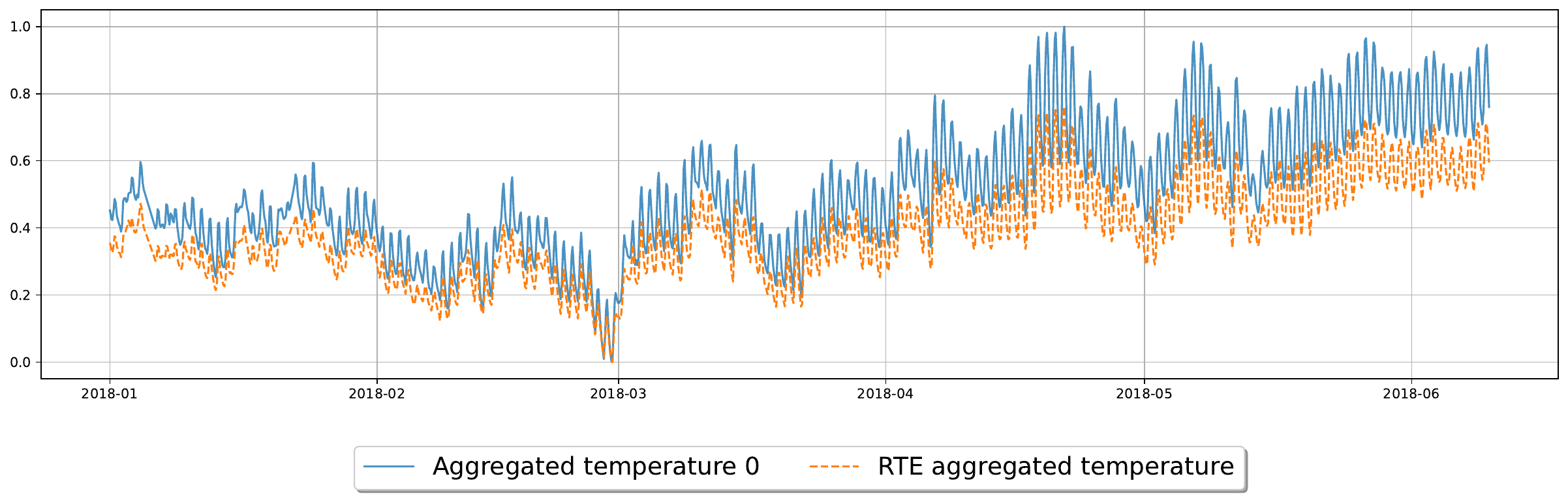}
         \caption{Dotted line: the temperature signal aggregated using weights from RTE, used in the GAM and EnergyDragon models. Solid line: the aggregated temperature signal found by the weather modeling module.}
         \label{fig:temp2}
     \end{subfigure}
     \hfill
     \begin{subfigure}[b]{0.95\textwidth}
         \centering
         \includegraphics[width=\textwidth]{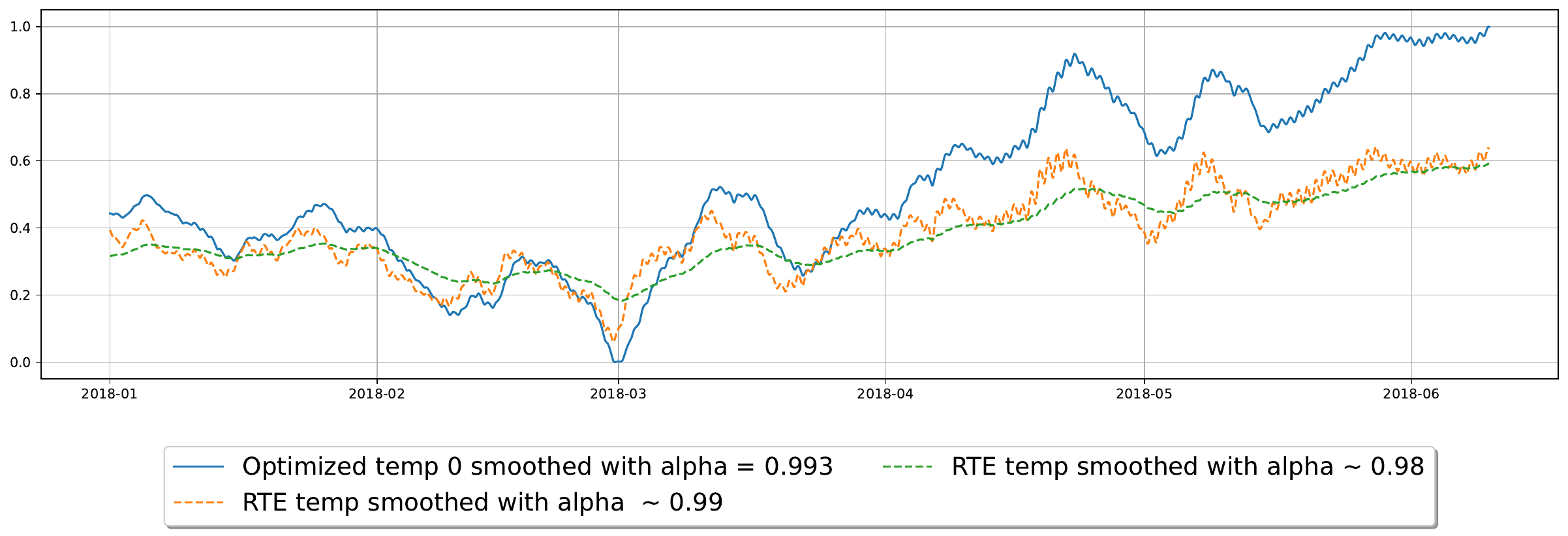}
         \caption{Dotted line: the smoothed temperature signals used in the GAM and EnergyDragon models. Solid line: the aggregated temperature signal smoothed by the weather modeling modeling module.}
         \label{fig:smoothed_temp2}
     \end{subfigure}
        \caption{Comparison between the weather as modeled within the GAM and EnergyDragon models, versus the weather modeled by the DNN based Weather Modeling module, for a model having a MAPE of 1.85\%}
        \label{fig:wm2}
\end{figure}

During the optimization phase, we noticed that the exponential smoothing layer exhibits superior performance in comparison to alternative modules. As Figure~\ref{fig:rnn_through_time} shows, there are rapidly no RNNs left in the new DNNs created. We noticed they are no longer employed after the 200th neural network is created (the initial population is 100 individuals, and ultimately, more than 2.000 models are evaluated during optimization). An examination of the output of the weather modeling module reveals the manner in which the DNN has modeled the weather. This modeled weather can then be compared to the data supplied by RTE, which was used in the GAMs and EnergyDragon models. Figures~\ref{fig:wm_1} and \ref{fig:wm2} show a comparison between the data as modeled by the functions given by RTE, and the one found by two DNNs using the Weather Modeling modules and achieving good performance (respectively $2.1\%$ and $1.85\%$ of MAPE). We can see that DNN's modeling close from that proposed by RTE, without being identical. It's interesting to notice, for example, that wind is almost identically represented, while for temperature we have two different aggregations. One is very close to the signal proposed by RTE, the other is opposite and larger in amplitude. As for smoothing, while Figure~\ref{fig:wm2} has a smoothing fairly close to that used in the GAM and EnergyDragon models, for Figure~\ref{fig:wm_1} the first smoothing coefficient is much lower.

\begin{figure}[htpb]
     \centering
     \begin{subfigure}[b]{0.45\textwidth}
         \centering
         \includegraphics[width=\textwidth]{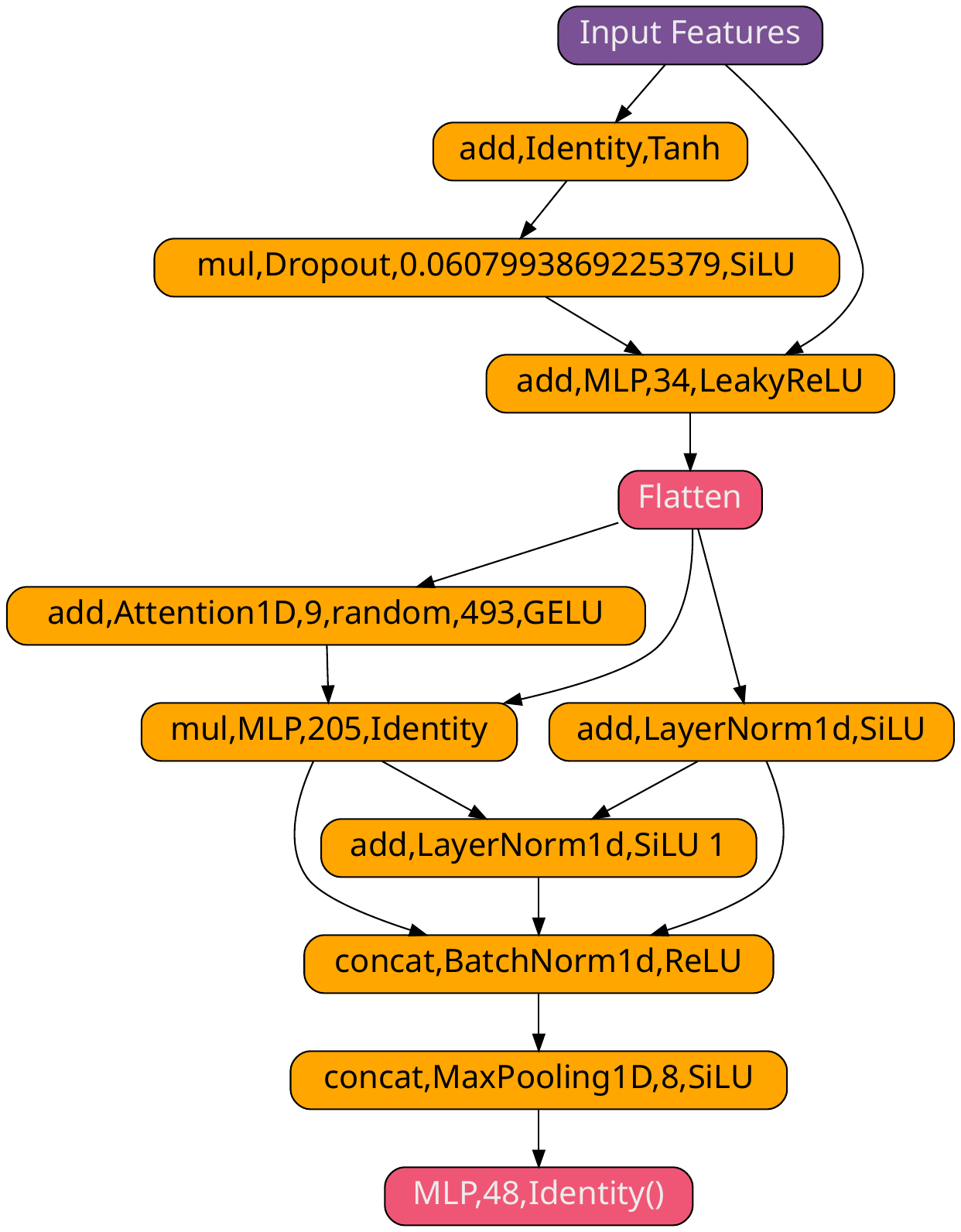}
         \caption{Best model found by EnergyDragon without the automated weather modeling part.}
         \label{fig:best_energy_dragon}
     \end{subfigure}
     \hfill
     \begin{subfigure}[b]{0.45\textwidth}
         \centering
         \includegraphics[width=\textwidth]{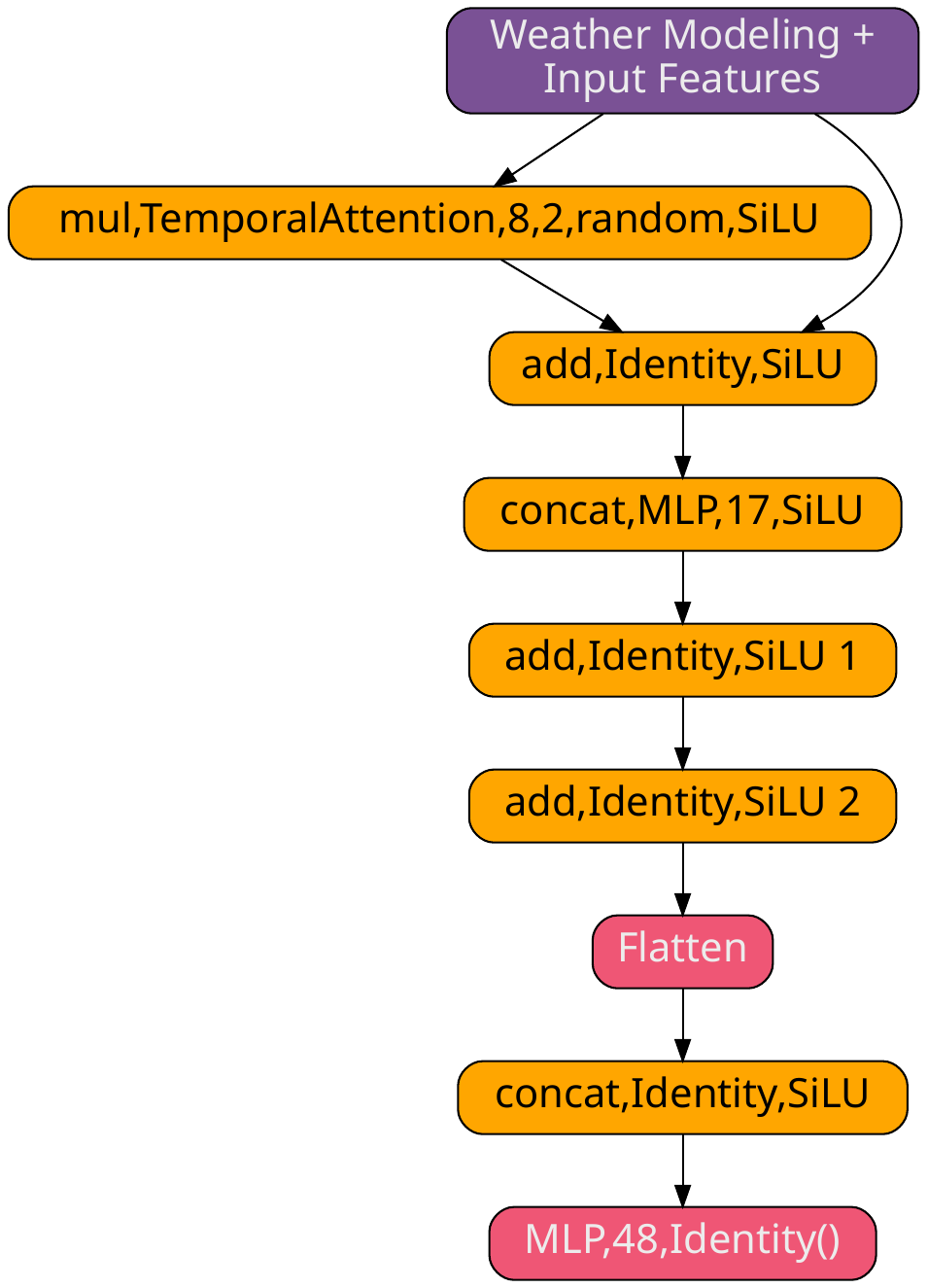}
         \caption{Best model found by EnergyDragon with the automated weather modeling part.}
         \label{fig:best_wm_energy_dragon}
     \end{subfigure}
        \caption{Best models found by EnergyDragon without and with the automated weather modeling module.}
        \label{fig:best_archis}
\end{figure}

Finally Figure~\ref{fig:best_archis} shows the models found by EnergyDragon with and without the weather modeling part. We focused on the core of the network without showing the weather modeling brick, to only compare the load forecasting network. The structure of EnergyDragon with the weather modeling network shown Figure~\ref{fig:best_wm_energy_dragon} is a lot simpler than the one without shown Figure~\ref{fig:best_energy_dragon}. It can be hypothesized that the weather is represented in a more comprehensible way for the DNN thanks to the weather modeling module. As a result, fewer transformations would be required to output the load forecast.

\section{Conclusion}\label{part:conclusion}

In conclusion, this article builds upon the work initiated by \citet{keisler2024automated} on automated deep learning for load forecasting. In this initial article, a framework, designated as EnergyDragon, was proposed for the optimization of both the architecture and hyperparameters of Deep Neural Networks, specifically designed for load forecasting. This article improves upon the previous work by incorporating an automated spatio-temporal weather modeling approach based on DNN and a recalibration module based on Kalman filtering. The efficacy of our approach is evaluated in a more dynamic and operational context, namely the national French load during the sobriety period. 

To automate the spatio-temporal representation of weather, we have maintained a close alignment with the functions employed in the state of the art for load forecasting. In the Section~\ref{part:experiments}, we demonstrate that the representations identified by our DNNs are close to those used in the other models from our baseling. This approach offers the advantage of remaining interpretable, enabling a comparison between the DNN-generated model and the insights derived from domain expertise. However, these preliminary results could probably be further enhanced by incorporating more sophisticated DNNs layers and employing over-parameterization.

\newpage
\bibliographystyle{abbrvnat}
\bibliography{references}

\end{document}